\pdfoutput=1

\documentclass[11pt]{article}

\usepackage[final]{acl}

\usepackage{times}
\usepackage{geometry}
\usepackage{tabularx}
\usepackage{latexsym}
\usepackage{amsmath}
\usepackage{amssymb}
\usepackage[T1]{fontenc}

\usepackage[utf8]{inputenc}

\usepackage{microtype}

\usepackage{inconsolata}

\usepackage{graphicx}

\title{Improvisational Games as a Benchmark for Social Intelligence of AI Agents: The Case of Connections}

\author{Gaurav Rajesh Parikh\textsuperscript{*} \\
  Department of Computer Science\\Duke University\\
  308 Research Drive\\
  Durham, NC 27708 \\
  \texttt{gaurav.rajesh.parikh@duke.edu} \\\And
  Angikar Ghosal\textsuperscript{*} \\
  Graduate School of Business\\Stanford University\\
  655 Knight Way\\
  Stanford, CA 94305 \\
  \texttt{angikar@stanford.edu} \\}

\begin{document}
\maketitle
\begin{abstract}
We formally introduce a improvisational wordplay game called Connections to explore reasoning capabilities of AI agents. Playing Connections combines skills in knowledge retrieval, summarization and awareness of cognitive states of other agents. We show how the game serves as a good benchmark for social intelligence abilities of language model based agents that go beyond the agents' own memory and deductive reasoning and also involve gauging the understanding capabilities of other agents. Finally, we show how through communication with other agents in a constrained environment, AI agents must demonstrate social awareness and intelligence in games involving collaboration. 
\end{abstract}
\section{Introduction}
In Season 6, Episode 4 of `The Big Bang Theory' \cite{bigbang_reentry_2012}, the characters Sheldon Cooper and Leonard Hofstadter lose a game of Pictionary to characters Amy Fowler and Penny, because Sheldon's clues being undecipherable to his teammates, despite the clues ``making sense'' to Sheldon himself, who is depicted as a very intelligent physics savant. Clearly, games that involved communication with team-members in a constrained fashion (not spelling out the word explicitly) require intelligence that go beyond one's own vocabulary and semantic connections, it involves understanding how to effectively communicate via shared world knowledge and mutual understanding. It is known \cite{zhang2020connecting} that both semantic categories and relations are represented by spatially overlapping cortical patterns, however the exact way in which people perceive semantic relations are highly variable \cite{chaffin1984similarity} and can depend on socio-cultural, educational and occupational factors. However, in such scenarios, understanding the ways in other agents perceive semantic relations and communicating accordingly is an important aspect of social intelligence.\\\\
We now demonstrate the game ``Connections'', a popular form of oral game-play with multiple variants played in different formats globally with no clear origin. Similar games that involve deductive reasoning along with gauging the mental states of other participants, including `Mafia', \cite{mafia_game} have been long popular as party games played by young adults. The game Connections requires the ability to satisfy structural game constraints and most importantly the awareness of capabilities of other agents for effective game-play. We now describe the rules of the game, and formalize it in later sections.\\\\
In Connections, one player (the Setter) decides on a word and other players(Guessers) collectively attempt to discover this word.  Initially, a single letter is revealed to players. Game play progresses when any player comes up with any clue that may be answered by the starting letter. The clue may be unrelated to the Setter's word but should be answered by a word with the same prefix. A connection occurs if another player is able to pick up the clue and simultaneously guess the word based on the clue without the setter blocking the clue by also guessing the word based on the clue correctly. Every time a guesser player is able to guess the clue by a different guesser player without being blocked, a letter is revealed. In each round, every player (apart from the one giving a clue) can make only one guess, including the setter. As more of the word is revealed over time, guessing words and making clues becomes more constrained. Ultimately the game should culminate with a final clue that the setter is unable to block (as it is the word the setter had in mind). This leads to the termination of the game. The guessers win if they together guess the word within a certain number of clues (e.g., $50$), else the setter wins.\\\\
Consider this example - three players sitting in a circle with Player 0 assigned as the Setter to start the game. The Setter thinks of a word and writes it down without showing the word to other players. 
The word is ``Catamaran''.
\begin{itemize}
    \item \textbf{Setter:} ``The first letter is \textbf{C}''
    \item \textbf{Player 1:} ``Garfield like?''
    \item \textbf{Setter:} ``Not a \textit{Cat}'' [Example of clue blocking]
    \item \textbf{Player 2:} ``Induces a pause in a sentence''

    \item \textbf{Player 1 and Player 2 (simultaneously):} ``\textbf{C}omma'' [Successful Connection]
    \item \textbf{Setter:} ``The next letter is \textbf{A}''
    \item \textbf{Player 1:} ``Language in Western Europe''
     \item \textbf{Player 1 (simultaneously):} ``\textbf{CA}talan'' 
    \item \textbf{Player 2 (simultaneously):} ``\textbf{CA}stillian''    [Guessor Wrong] 
    \item \textbf{Player 1:} ``Rug, sometimes flying''
 
    \item \textbf{Player 1 and Player 2 (simultaneously):} ``\textbf{CA}rpet'' [Successful Connection]
    
    \item \textbf{Setter:} ``The next letter is \textbf{T}''
 
    \item \textbf{Player 2:} ``Baby butterfly''
  
    \item \textbf{Player 1 and Player 2 (simultaneously):} ``\textbf{CAT}erpillar''[Successful Connection]
    \item \textbf{Setter:} ``The next letter is \textbf{A}''
    \item \textbf{Player 1:} ``Speeds a chemical reaction''

    \item \textbf{Player 1 and Player 2 (simultaneously):} ``\textbf{CATA}lyst''[Successful Connection]
    \item \textbf{Setter:} ``The next letter is \textbf{M}''
    \item \textbf{Player 1:} ``Type of boat"

    \item \textbf{Player 1 and Player 2 (simultaneously):} ``\textbf{CATAMARAN}'' [End of game play with revelation of the word]
\end{itemize}



 Normally played by a group of people, we showcase the effects of game-play by LLM agents to explore cognition, contextual understanding, and reasoning capabilities of large language models. A wide range of benchmarks have been proposed for measuring the capabilities of Large Language Models \cite{chang2024survey}. Summarization \citep{ji2023survey,zhang2024benchmarking,pu2023chatgpt}
 question answering\citep{jiang2021can}, subject matter expertise and test taking, as well as the understanding of social knowledge and social cues\citep{choi-etal-2023-llms}. However, there is limited work on the ability of Large Language Models to show awareness to reason about the abilities of other players and adapt. 
 
 Playing a game like Connections requires a variety of different reasoning capabilities, requiring both the ability to summarize, respond to questions, have both general and subject knowledge expertise. Most important, however, is the ability of the agent to interact with and gauge the ability of other game players. Not only is the player trying to come up with a clever clue, there is a delicate balance between coming up with clues that are easy enough that other players may get them, yet tricky enough so that the Setter does not block the clue. Beyond just obtaining clues and answers via lookup in clue-answer databases, agents must be capable of generating clues taking into account the knowledge and ability of other players.
 
  Our goal in this work is to showcase that LLM agents have the capacity to play such a game and reason about the abilities of other players in order to successfully collaborate.

\section{Formal Description}
 Consider a vocabulary of words $\mathfrak{U}$ with words $\mathfrak{u}_i\in\mathfrak{U}$. The words are made of letters chosen from an alphabet set $\aleph$ with size $L$, with characters $\mathsf{c}_1, \mathsf{c}_2, \dots, \mathsf{c}_L$. Word $\mathfrak{u}_i$ has length $\mathfrak{t}_i$, with characters $\mathsf{s}_{i1}, \mathsf{s}_{i2},\dots \mathsf{s}_{i\mathfrak{t}_i}$. For simplicity, we use the English alphabet with length $26$. \\\\
 Consider a set of $n+1$ players, with indices $0,1,\dots, n$ where $n\geq 2$. Player $0$ is denoted as the \textbf{Setter}. Let us focus on the simpler game first (word is chosen and then fixed).\\\\
 Different players have different working vocabularies that they can think of candidate words from. Let the working vocabularies for player $j$ be $\mathfrak{U}_j$, where all $\mathfrak{U}_j\subseteq \mathfrak{U}$. Consider a single play in a run of the game when a player $j$ thinks of a word $\mathsf{w}$ agreeing with the revealed words and thinks of a particular clue $\texttt{p}$. The clue should be such that the clue is "associated" with the word (e.g., $\texttt{domesticated animal}\rightarrow \texttt{cat}$). We represent the semantic embedding representations of each player $j$ through a function $\Phi_j$ that maps a word or a string clue to a vector of $m$ dimensions. Hence, we assume that the embedding for word $\mathfrak{w}$ is one of the "closest" words to the embedding representation for the clue $\texttt{p}$. Formally, for small integer $k$, and a given clue phrase $\mathsf{p}$, $$\mathsf{w}\in\underset{\mathfrak{u}}{\text{arg-}k\text{-max}}(\Phi_j(\mathfrak{u})\cdot\Phi_j(\texttt{p}))$$ where the geometric proximity of the embedding representation is just the vector dot product. Ideally, $k=1$ (the word in mind is the closest word to the clue).\\\\Note that even if the players know the same word, the exact semantic connections between this word and other words and phrases could be different, so we assume each player has their own representation $\Phi_j$. But for words that are most semantically close to the given clue in the mental representation for one player, they should be similarly close to the given clue in the mental representation for other players, i.e., different players can have different perceptions of the relative (lack of) semantic closeness of \texttt{drinking container} and \texttt{chair}, but they should have similar perception of the semantic closeness of \texttt{drinking container} and \texttt{cup}, even though the "closest" word to the clue \texttt{drinking container} may be \texttt{chalice} to some other player. More formally, we can assume that, if for player $j$ and a clue $\texttt{p}$, $$\mathsf{w}\in\underset{\mathfrak{u}}{\text{arg-}k\text{-max}}(\Phi_j(\mathfrak{u})\cdot\Phi_j(\texttt{p}))$$then for other $j'$, there exists a constant $\epsilon$ such that $$(1-\epsilon)(\Phi_j(\mathsf{w})\cdot\Phi_j(\texttt{p}))$$$$\leq(\Phi_{j'}(\mathsf{w})\cdot\Phi_{j'}(\texttt{p}))$$$$\leq(1+\epsilon)(\Phi_j(\mathsf{w})\cdot\Phi_j(\texttt{p}))$$
 This is why this gameplay is a valuable benchmark to explain the effect of incorporation of socio-cultural-educational aspects of an agent - different human players have their unique learning histories and are familiar with different sets of vocabularies. In a real human gameplay, clues do involve shared personal information that the setter does not know (such as \texttt{Sport I played in high school}, when \texttt{B} is revealed, where the setter does not but some other guesser knows the answer is \texttt{badminton}, despite \texttt{baseball} or \texttt{basketball} being more `obvious' answers.\\\\
 An ideal clue that is revealed by a player should not be too vague nor too obvious. A clue like \texttt{capital of France} will be immediately blocked by the Setter as \texttt{Paris} (unless \texttt{Paris} is the answer) and a clue like \texttt{a particular animal} is highly likely to invoke different animals as the "closest" word to the given clue in different players' mental representations. Hence, formally if the player $j$ thinks of a word $\mathsf{w}$ and clue $\texttt{p}$, then $$\lambda_L<\Phi_j(\mathsf{w})\cdot\Phi_j(\texttt{p})<\lambda_U$$ (clue is not too obvious, but not too unrelated to the word in mind). If $k>1$, for $\mathsf{u}\neq \mathsf{w}, \mathsf{u}\in\underset{\mathfrak{u}}{\text{arg-}k\text{-max}}(\Phi_j(\mathfrak{u})\cdot\Phi_j(\texttt{p}))$, other words that are sufficiently close to the clue, $$\Phi_j(\mathsf{u})\cdot\Phi_j(\texttt{p}))<\lambda_U$$ as well (the other players should not guess an incorrect connection)!\\\\
 For word $\mathfrak{u}_i$, once letters $1,2,\dots, k$ have been revealed, future words which clues need to be generated for need to have $\mathsf{s}_{i1}\mathsf{s}_{i2}\dots\mathsf{s}_{ik}$ as a prefix. This constrains the sample space in which new words are generated probabilistically. We obtain a probabilistic explanation of the heuristic to generate a clue for a word, assuming that the AI agents are uniform (i.e., the guesser who comes up with a clue has no additional information regarding the semantic alignment of other agents). Given a partially revealed word, assume that the guesser is agnostic as to which word in their vocabulary adhering to the constraints, they actually choose to give a clue for. Then, given a clue for this word, let the probability of a given agent guessing it correctly is $p$. Note that every agent gets to guess once in a given round/turn. We want the setter to not guess correctly (probability $1-p$) and at least one of the $n-1$ other guessers to guess correctly. Hence, the probability of success in this run (a new letter is revealed) is $(1-p)(1-(1-p)^{n-1})$. Maximizing this, we see that we want $$p^*=1-\left(\frac{1}{n}\right)^{\frac{1}{n-1}}$$ For $n=2,3,4,5$, we have $p^*\approx 0.5, 0.43, 0.37, 0.33$. More the players involved, the clue giver can give a clue somewhat vaguer as decreasing the odds of the setter blocking the word is overcome by the odds of some guesser getting it. The procedure of generating a suitable clue thus has a mental model where the degree of semantic overlap/connection between the clue and the word (i.e., the dot product) can be monotonically mapped to the probability of success for the guessers overall. Clue generation needs a mental proxy to find a roughly optimal clue.\\\\
We propose that such wordplay games are important for another social intelligence ability of AI agents - \textit{the ability to gauge a person's background over runs}. This is an important step, we opine, in making AI have intelligence abilities of humans that go beyond deductive reasoning, for the act of "finding" an optimal clue and an optimal word assuming uniform agents is a mathematical function. However, over arbitrarily many turns of the game, an agent should be able to detect which words some agents are more likely to get connections with and less likely with. For example, over turns, it is possible for humans to gauge who has a better grasp of TV shows and who has a better grasp of medical terminology, and choose accordingly. Let the working vocabularies for player $i$ be $\mathfrak{U}_i$. Let $\mathfrak{U}^\dag=(\cup_{i=1}^n\mathfrak{U}_i)\setminus \mathfrak{U}_0$ be the vocabulary that is common knowledge among the non-setter players but unknown to the setter. A good clue targets $\mathfrak{U}^\dag$.\\\\We can also describe the word and clue generation procedure probabilistically. In \cite{arora2016latent}, the authors build on \cite{mnih2007three} and describe a generative model that treats corpus generation according to a log-linear production model where the probability of generation of a word is proportional to the exponential of the dot product with the discourse vector, which is undergoing a slow random walk and $v_w$ is the embedding vector of word $w$. Analogously, consider the "true" discourse vector for every agent $j$ as $d^*_j$. If the agent knew the discourse vectors for every other agent, then the word and clue generation procedure could be "aligned" towards the discourse vector average for guesser agents and away from the discourse vector average for setter agents by considering the truncated probability distribution (choosing the top $k$ probability words), where $v_w$ is generated by the $\Phi_i$ mapping as mentioned before. $$\mathbf{P}[\text{word }w]\propto \exp(\langle v_w, d_{\text{setter:avg}}\rangle-\langle v_w, d_0\rangle)$$ But, when an agent encounters a new agent, the social intelligence of gauging the cognitive abilities of the new agent is as follows:
\begin{itemize}
    \item Let $d_{i\leftarrow j}$ be the discourse vector of agent $j$ as perceived by agent $i$. For all $j$ and a fixed $i$, let this perceived discourse vector be set to $d^\dag$, which is the discourse vector corresponding to "common knowledge", the basic ideas people know about.
    \item A word and the clue is generated by the generative process. Initially, $d_{i\leftarrow\text{setter:avg}}$ and $d_{i\leftarrow 0}$ are both set to $d^\dag$, so the probability of generating a word is uniform, i.e., "any" random word and clue is generated.
    \item Let $\eta$ be a suitably small constant. Each word $w$ has its corresponding embedding vector $v_w$. If agent $j$ fails to guess the clue for the word, we subtract $\eta v_w$ from the current value of $d_{i\leftarrow j}$, and if agent $j$ correctly guesses the clue for the word, we add $\eta v_w$ to the current value of $d_{i\leftarrow j}$. $d_{i\leftarrow\text{setter:avg}}$ is recalibrated accordingly.
    \item This gradient descent/ascent-like procedure mimics the procedure of understanding the cognitive backgrounds of other agents using social intelligence, as it allows future turns to generate clues more likely to be understood by guessers and less so by the setter by updating the perceived values of $d_{i\leftarrow j}$.
    \item Likewise, when one guesser player $i$ tries a clue for a word $w$, other guesser players, at the end of the turn, add $\eta v_w$ to the perceived discourse vector $d_{j\leftarrow i}$, as guesser $i$ must have their own discourse vector more aligned with the word $w$.
\end{itemize}
\section{Connections}
\subsection{Semantic Networks}
The key observation is that games utilizing semantic networks do not have a strict notion of monotonicity - i.e., it is not that one player agent strictly \textit{dominates} another player agent by correctly guessing the word to a clue whenever the other agent guesses it correct, or a notion of probabilistic monotonicity, i.e., it is not that one player agent has strictly higher probability of correctly guessing the word to a clue compared to another agent. Similar to negotiation games, \cite{davidson2023evaluating}, "powerful" agents can lose to weaker ones. \\\\
Semantic networks serve as the representational basis of our cognitive system as prominent models of memory and reasoning.  Semantic networks represent knowledge through relations between abstract objects \citep{borge2010semantic} and could vary considerably between individuals and with different life experiences\citep{benedek2017semantic,dubossarsky2017quantifying,morais2013mapping}
The goal of a game like Connections is to elucidate, through an iterative process, the discovery of sections of the semantic networks of diverse players that are structurally similar by the task of proposing a set of valid clues that should elicit the same response that must also vary from that of the Setter.  Using LLMs for game play requires semantic priming through a prompt to identify the specific role of the agent since the base model for each agent may be the same. 
\subsubsection{ Knowledge-Based Clues}
Semantic clues would most often rely on having a shared knowledge about a certain concept or idea. Such clues should be effective to identify the target word precisely if other players have the requisite knowledge. For instance the clue ``Speeds a chemical reaction" is precipitated on some other player having some knowledge of chemical reactions. The domain and technical depth of clues produced is linked to the depth and breadth of vocabulary that players may share\citep{vermeer2001breadth}. In the human player setting, if two players are able to discover a shared area of knowledge which the Setter is not familiar with the game is significantly simplified. The players can now use this to produce clues that will not be blocked, given they may have a sufficiently rich vocabulary to be able to play the game as the prefix elongates.
\subsubsection{Personal Clues}
A very effective means of giving clues is those that are built on shared experiences of players or based on some level of knowledge that players have about other players. Such clues are expected to largely be successful as they may not hold any meaning to the Setter or even other players but would allow those who have the appropriate context to guess a word effectively. However, as the space of words gets restricted as more letters are revealed, it is unlikely that there might be adequate personal clues that exist. I.e. forming a personal clue to the prefix ``C" might be easy as `` What pet I have" but not to the prefix ``CATA".
\subsubsection{Word Association Clues}
Words have associative links to each other to varying degrees, both semantically, structurally as well in the cognitive representation in memory \citep{karwoski1948psychological}, \citep{kent1910study}. Word associative clues may be used in the game such as ``Good-Bad",  ``Hero-Villain", ``Sun-Moon" that leverage this associative structure. Often words exist where the most frequent response to a given clue is several times as frequent as the next-most-frequent response \citep{woodworth1954experimental} or there are well developed word association norms that have \citep{palermo1965changes, nelson2004university,toglia1978handbook}. A key challenge in playing the game with associative clues is that though such clues elicit the correct response it is likely the Setter will easily be able to block such clues. 
\section{Experiments}
\begin{table*}[!ht]
\centering
\small 
\setlength{\tabcolsep}{4pt} 
\begin{tabular}{|>{\raggedright\arraybackslash}p{2.5cm}|r|r|r|r|}
\hline
\textbf{Word} & \textbf{Reveals} & \textbf{Guesser Wrong} & \textbf{Setter Blocked} & \textbf{Iterations} \\ \hline
kaleidoscope  & 0                & 1                     & 7                      & 8              \\ \hline
xenophobia    & 1                & 2                     & 4                      & 7              \\ \hline
labyrinthine  & 9                & 26                    & 12                     & 47             \\ \hline
uppercases    & 8                & 20                    & 18                     & 46             \\ \hline
entrepreneur  & 4                & 12                    & 26                     & 42             \\ \hline
laboriously   & 5                & 21                    & 23                     & 49             \\ \hline
encyclopedia  & 5                & 17                    & 38                     & 60             \\ \hline
villeinage    & 4                & 30                    & 40                     & 74             \\ \hline
exploration   & 5                & 54                    & 22                     & 81             \\ \hline
photosystem   & 6                & 31                    & 48                     & 85             \\ \hline
elaborately   & 7                & 48                    & 35                     & 90             \\ \hline
goldfish      & 6                & 42                    & 53                     & 101            \\ \hline
revolving     & 6                & 58                    & 37                     & 101            \\ \hline
precaution    & 5                & 54                    & 47                     & 106            \\ \hline
multinomial   & 4                & 45                    & 59                     & 108            \\ \hline
precipitate   & 9                & 39                    & 68                     & 116            \\ \hline
metamorphosis & 7                & 39                    & 71                     & 117            \\ \hline
circumvented  & 9                & 83                    & 72                     & 164            \\ \hline
conjunction   & 5                & 123                   & 45                     & 173            \\ \hline
\end{tabular}
\caption{Comparisons Across Different Words, Two Guessers, One Setter (Ordered by Iterations)}
\label{tab:acl_table}
\end{table*}

We run experiments with three players using the \textbf{GPT-4o} model as the language model powering the reasoning capabilities of the players. We assign the role of the Setter to Player 0, and Player 1 and Player 2 play the roles of the Guesser. Initially, Player 0, decides a word and reveals only the first letter to players after which the game play proceeds as described. The results of running this game play over a few iterations are recorded in Table \ref{tab:acl_table}. The `Reveals' column describes the number of successful letter reveals that happened before the word was guessed, while the  `Guesser wrong' and `Setter Blocked' columns respectively describe the cases where either the other player guessed the response to the clue incorrectly or the clue was blocked by the setter. \\\\
 We illustrate a sample of the full interaction between agents in Appendix \ref{sample_gameplay}. We used a variety of prompts to guide the actions of the LLM agents which are detailed in Appendix \ref{prompts}.
\subsection{Reasoning Capabilities}
\begin{figure}
    \centering
    \includegraphics[scale=0.4]{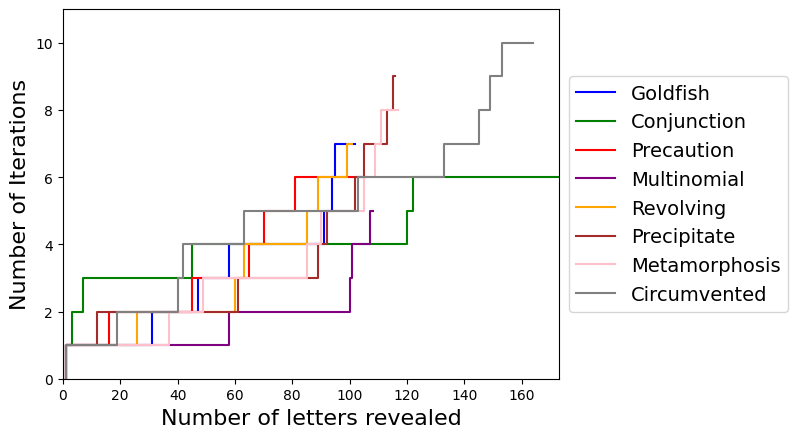}
    \caption{Number of Characters Revealed over iterations for words with more than 100 iterations of game play }
    \label{fig:iterations_revelations}
\end{figure}

We observe that players showcase interesting reasoning capabilities. Certain behaviors are well expected and similar to human agents such as the relatively smaller number of iterations observed for the word starting with `X' as opposed to a much higher number of iterations of game play for words starting with `C'. Though there is a high degree of chance in guessing the right word, the set of word beginning with `X' is much smaller than with `C'. \ref{fig:iterations_revelations} showcases the number of letters revealed over the number of iterations before the word is correctly guessed. We observe that models generally tend to need fewer iterations beyond the first few letter guesses as the constrained space of words given a longer prefix is smaller, however, an interesting non-human behaviour that is observed is that sometimes there is a lot of iterations between the final letter being revealed and the word being guessed often as a result of the model making bizarrely uncommon guesses prior to make guesses that are far more obvious. 
\subsection{Social Awareness}
In standard game play, we observe that an overwhelming number of clues are blocked by the Setter, since it is likely given that we are using the same language model, the semantic network to represent words for all agents is quite similar. However, by introducing in-context learning\citep{lampinen2022can}, we can instigate varying behaviours. Notably, by priming the agent on aspects such as profession, cultural context, age, we can instigate a change in the semantic network  of the model\citep{benedek2017semantic,dubossarsky2017quantifying,morais2013mapping}. Under such a framework, we observe the AI agents' ability to utilize this context in the game play through the use of concerted vocabulary and clues if agents are explicitly made aware of the the priming for other agents. However, unlike humans, AI agents may be unable to gauge this context, or may fail to act in a way to discover this information through the use of diverse clues to attempt to find regions of the semantic network that might be similar between agents without any prompting.
\section{Future Work}
Our work considers a simple game where the word picked by the Setter is fixed. A more sophisticated version of this game, and one that is common in game-play is where the word is not necessarily fixed (not written down before) but has to agree with the game played so far. For example, the Setter in this particular game-play may realize after the letters `C', `A',`T' and `A' being revealed that once the letter `M' is revealed, there are not enough words besides the chosen word `catamaran' (and the Setter can no longer block it). However, if the Setter now switches to `cataclysm', the Setter has an advantage because there are other words starting with `catac', such as `catacomb'. The Setter wants to pick a word that is sufficiently obscure but prefixes do not give the word away easily. Modifying our experiments to not have the word be fixed \textit{a priori} has limited success as the LLM is not able to robustly recall the game history to make improvisational decisions. Future work can explore the proof-of-concept alignment of perceived discourse vectors of other agents, given AI agents with a variety of historical training/life experiences that goes beyond priming. A LLM agent endowed with social intelligence capabilities shows a first step towards more complex reasoning patterns that require sensory and cognitive abilities as well as the power of agency to autonomously adapt to changing scenarios. \cite{liu2017lifelong}
\section*{Acknowledgments}
The authors wish to confirm that there are no known conflicts of interest associated with this publication and there has been no significant financial support for this work that could have influenced its outcome. The authors wish to confirm that there are no known ethical concerns associated with this publication.
\bibliography{custom}

@misc{bigbang_reentry_2012,
  title = {The Re-Entry Minimization},
  author = {Chuck Lorre and Bill Prady},
  year = {2012},
  note = {Season 6, Episode 4},
  howpublished = {Television series episode},
  series = {The Big Bang Theory},
  network = {CBS},
}

@article{liu2017lifelong,
  title={Lifelong machine learning: a paradigm for continuous learning},
  author={Liu, Bing},
  journal={Frontiers of Computer Science},
  volume={11},
  number={3},
  pages={359--361},
  year={2017},
  publisher={Springer Nature BV}
}

@inproceedings{davidson2023evaluating,
  title={Evaluating Language Models Through Negotiations},
  author={Davidson, Tim Ruben and Veselovsky, Veniamin and Kosinski, Michal and West, Robert},
  booktitle={The Twelfth International Conference on Learning Representations},
  year={2023}
}

@inproceedings{mnih2007three,
  title={Three new graphical models for statistical language modelling},
  author={Mnih, Andriy and Hinton, Geoffrey},
  booktitle={Proceedings of the 24th international conference on Machine learning},
  pages={641--648},
  year={2007}
}

@article{arora2016latent,
  title={A latent variable model approach to pmi-based word embeddings},
  author={Arora, Sanjeev and Li, Yuanzhi and Liang, Yingyu and Ma, Tengyu and Risteski, Andrej},
  journal={Transactions of the Association for Computational Linguistics},
  volume={4},
  pages={385--399},
  year={2016},
  publisher={MIT Press One Rogers Street, Cambridge, MA 02142-1209, USA journals-info~…}
}

@article{ji2023survey,
  title={Survey of hallucination in natural language generation},
  author={Ji, Ziwei and Lee, Nayeon and Frieske, Rita and Yu, Tiezheng and Su, Dan and Xu, Yan and Ishii, Etsuko and Bang, Ye Jin and Madotto, Andrea and Fung, Pascale},
  journal={ACM Computing Surveys},
  volume={55},
  number={12},
  pages={1--38},
  year={2023},
  publisher={ACM New York, NY}
}

@article{lampinen2022can,
  title={Can language models learn from explanations in context?},
  author={Lampinen, Andrew K and Dasgupta, Ishita and Chan, Stephanie CY and Matthewson, Kory and Tessler, Michael Henry and Creswell, Antonia and McClelland, James L and Wang, Jane X and Hill, Felix},
  journal={arXiv preprint arXiv:2204.02329},
  year={2022}
}

@inproceedings{choi-etal-2023-llms,
    title = "Do {LLM}s Understand Social Knowledge? Evaluating the Sociability of Large Language Models with {S}oc{KET} Benchmark",
    author = "Choi, Minje  and
      Pei, Jiaxin  and
      Kumar, Sagar  and
      Shu, Chang  and
      Jurgens, David",
    editor = "Bouamor, Houda  and
      Pino, Juan  and
      Bali, Kalika",
    booktitle = "Proceedings of the 2023 Conference on Empirical Methods in Natural Language Processing",
    month = dec,
    year = "2023",
    address = "Singapore",
    publisher = "Association for Computational Linguistics",
    url = "https://aclanthology.org/2023.emnlp-main.699",
    doi = "10.18653/v1/2023.emnlp-main.699",
    pages = "11370--11403",
}

@article{jiang2021can,
  title={How can we know when language models know? on the calibration of language models for question answering},
  author={Jiang, Zhengbao and Araki, Jun and Ding, Haibo and Neubig, Graham},
  journal={Transactions of the Association for Computational Linguistics},
  volume={9},
  pages={962--977},
  year={2021},
  publisher={MIT Press One Rogers Street, Cambridge, MA 02142-1209, USA journals-info~…}
}

@article{pu2023chatgpt,
  title={ChatGPT vs human-authored text: Insights into controllable text summarization and sentence style transfer},
  author={Pu, Dongqi and Demberg, Vera},
  journal={arXiv preprint arXiv:2306.07799},
  year={2023}
}

@article{chang2024survey,
  title={A survey on evaluation of large language models},
  author={Chang, Yupeng and Wang, Xu and Wang, Jindong and Wu, Yuan and Yang, Linyi and Zhu, Kaijie and Chen, Hao and Yi, Xiaoyuan and Wang, Cunxiang and Wang, Yidong and others},
  journal={ACM Transactions on Intelligent Systems and Technology},
  volume={15},
  number={3},
  pages={1--45},
  year={2024},
  publisher={ACM New York, NY}
}

@article{zhang2024benchmarking,
  title={Benchmarking large language models for news summarization},
  author={Zhang, Tianyi and Ladhak, Faisal and Durmus, Esin and Liang, Percy and McKeown, Kathleen and Hashimoto, Tatsunori B},
  journal={Transactions of the Association for Computational Linguistics},
  volume={12},
  pages={39--57},
  year={2024},
  publisher={MIT Press One Broadway, 12th Floor, Cambridge, Massachusetts 02142, USA~…}
}

@article{chaffin1984similarity,
  title={The similarity and diversity of semantic relations},
  author={Chaffin, Roger and Herrmann, Douglas J},
  journal={Memory \& Cognition},
  volume={12},
  pages={134--141},
  year={1984},
  publisher={Springer}
}

@misc{mafia_game,
  author = {Dimitry Davidoff and Andrew Plotkin},
  title = {Mafia},
  note = {Also known as Werewolf},
  howpublished = {\url{https://escaleajeux.fr/?principal=/jeu/mafid}},
  year = {n.d.},
}

@article{zhang2020connecting,
  title={Connecting concepts in the brain by mapping cortical representations of semantic relations},
  author={Zhang, Yizhen and Han, Kuan and Worth, Robert and Liu, Zhongming},
  journal={Nature communications},
  volume={11},
  number={1},
  pages={1877},
  year={2020},
  publisher={Nature Publishing Group UK London}
}

@article{borge2010semantic,
  title={Semantic networks: Structure and dynamics},
  author={Borge-Holthoefer, Javier and Arenas, Alex},
  journal={Entropy},
  volume={12},
  number={5},
  pages={1264--1302},
  year={2010},
  publisher={MDPI}
}

@article{benedek2017semantic,
  title={How semantic memory structure and intelligence contribute to creative thought: A network science approach},
  author={Benedek, Mathias and Kenett, Yoed N and Umdasch, Konstantin and Anaki, David and Faust, Miriam and Neubauer, Aljoscha C},
  journal={Thinking \& Reasoning},
  volume={23},
  number={2},
  pages={158--183},
  year={2017},
  publisher={Taylor \& Francis}
}

@book{toglia1978handbook,
  title={Handbook of semantic word norms.},
  author={Toglia, Michael P and Battig, William F},
  year={1978},
  publisher={Lawrence Erlbaum}
}

@article{vermeer2001breadth,
  title={Breadth and depth of vocabulary in relation to L1/L2 acquisition and frequency of input},
  author={Vermeer, Anne},
  journal={Applied psycholinguistics},
  volume={22},
  number={2},
  pages={217--234},
  year={2001},
  publisher={Cambridge University Press}
}

@article{nelson2004university,
  title={The University of South Florida free association, rhyme, and word fragment norms},
  author={Nelson, Douglas L and McEvoy, Cathy L and Schreiber, Thomas A},
  journal={Behavior Research Methods, Instruments, \& Computers},
  volume={36},
  number={3},
  pages={402--407},
  year={2004},
  publisher={Springer}
}

@article{palermo1965changes,
  title={Changes in the word associations of fourth-and fifth-grade children from 1916 to 1961},
  author={Palermo, David S and Jenkins, James J},
  journal={Journal of verbal learning and verbal behavior},
  volume={4},
  number={3},
  pages={180--187},
  year={1965},
  publisher={Elsevier}
}

@article{dubossarsky2017quantifying,
  title={Quantifying the structure of free association networks across the life span.},
  author={Dubossarsky, Haim and De Deyne, Simon and Hills, Thomas T},
  journal={Developmental psychology},
  volume={53},
  number={8},
  pages={1560},
  year={2017},
  publisher={American Psychological Association}
}

@article{morais2013mapping,
  title={Mapping the structure of semantic memory},
  author={Morais, Ana Sofia and Olsson, Henrik and Schooler, Lael J},
  journal={Cognitive science},
  volume={37},
  number={1},
  pages={125--145},
  year={2013},
  publisher={Wiley Online Library}
}

@article{karwoski1948psychological,
  title={Psychological studies in semantics: III. Reaction times for similarity and difference},
  author={Karwoski, Theodore F and Schachter, J},
  journal={The Journal of Social Psychology},
  volume={28},
  number={1},
  pages={103--120},
  year={1948},
  publisher={Taylor \& Francis}
}

@article{kent1910study,
  title={A study of association in insanity},
  author={Kent, Grace Helen and Rosanoff, Aaron Joshua},
  journal={American Journal of Psychiatry},
  volume={67},
  number={1},
  pages={37--96},
  year={1910},
  publisher={Am Psychiatric Assoc}
}

@book{woodworth1954experimental,
  title={Experimental psychology},
  author={Woodworth, Robert Sessions and Schlosberg, Harold},
  year={1954},
  publisher={Oxford and IBH Publishing}
}
\appendix
\section{Sample Gameplay}
\label{sample_gameplay}
Word chosen: 'XENOPHOBIA'\\
\texttt{\\
Setter :  X \\\\
Attempt Number  1 \\
Guesser 1 (mentally) :  Xylograph \\
Guesser 1 :  Woodblock printing technique \\
Setter :  Xylography \\
Guesser 2 :  Xylography \\
Setter attempted  XYLOGRAPHY , Player attempted  XYLOGRAPHY \\
Guesser Wrong! \\\\
Attempt Number  2 \\
Guesser 1 (mentally):  Xanthophyll \\
Guesser 1 :  Leaf pigment category \\
Setter :  Xanthophyll \\
Setter attempted  XANTHOPHYLL\\
Setter Blocked! \\
BLOCKED:  ['XANTHOPHYLL'] \\\\
Attempt Number  3 \\
Guesser 1 (mentally):  Xenoglossy \\
Guesser 1 :  Mysterious language phenomenon \\
Setter :  Xenoglossy \\
Setter attempted  XENOGLOSSY \\
Setter Blocked! \\
BLOCKED:  ['XANTHOPHYLL', 'XENOGLOSSY'] \\\\
Attempt Number  4 \\
Guesser 1 (mentally):  Xiphophyllous \\
Guesser 1 :  Sword-shaped leaves \\
Setter :  Xiphoid \\
Guesser 2 :  Xiphophyllous \\
Setter attempted  XIPHOID , Player attempted  XIPHOPHYLLOUS \\
Success! \\
Comparing  XIPHOPHYLLOUS  with  XENOPHOBIA \\
Game not over yet\\\\
Setter :  XE \\\\
Attempt Number  5 \\
Guesser 1 (mentally):  Xenolithic \\
Guesser 1 :  Foreign rock inclusion \\
Setter :  Xenolith \\
Guesser 2 :  Xenolith \\
Setter attempted  XENOLITH , Player attempted  XENOLITH \\
Guesser Wrong! \\\\
Attempt Number  6 \\
Guesser 1 (mentally):  Xenogenesis \\
Guesser 1 :  Alien life formation \\
Setter :  Xenogenesis \\
Setter attempted  XENOGENESIS \\
Setter Blocked! \\
BLOCKED:  ['XENOGLOSSY', 'XENOGENESIS'] \\\\
Attempt Number  7 \\
Guesser 1 (mentally):  Xerophthalmia \\
Guesser 1 :  Dry eye condition \\
Setter :  Xerophthalmia
Setter attempted  XEROPHTHALMIA \\
Setter Blocked! \\
BLOCKED:  ['XENOGLOSSY', 'XENOGENESIS', 'XEROPHTHALMIA'] \\\\
Attempt Number  8 
Guesser 1 (mentally):  Xenophobia \\
Guesser 1 :  Fear of foreigners \\
Setter (mentally):  Xenophobia (Cannot block anymore)\\
Guesser 2 :  Xenophobia \\
Player attempted  XENOPHOBIA \\
Success! \\
Comparing  XENOPHOBIA  with  XENOPHOBIA \\
Guessed Correctly. Game Over. \\
Reveals, GuesserWrong, SetterBlocked: \\ 1 ,  2 ,  4 \\
7 \\
}
\section{Prompts}
\label{prompts}
\begin{itemize}
    \item Prompt to generate a new word: \\\\\texttt{You are playing a wordplay game, where you are the setter player playing against 2 guesser players. You have to pick a legitimate English word of some suitable length. The guesser players have to guess the word gradually. For now, pick a totally random word and just output that word without any leading phrases. Do not output anything else, only one single word.}
    \item Prompt explaining the rules to the setter:\\\\\texttt{Here are the rules of the game. Initially, you reveal the first letter to all guesser players. In every round, one guesser player will come up with a suitable clue phrase whose answer begins with your revealed letter(s). If you can guess the answer to their clue and it is not the same as the word you had picked, then you will output in that round the answer to their clue. If your guess is the same as the answer to the clue a guesser player came up with, then there is a new round. If your guess is different than the answer to the clue a guesser player came up with, a different guesser player will try to guess the clue word. If this different guesser player guesses the answer correctly, then you will have to reveal the next character of the word. In future rounds, the other guesser players have to come up with clues whose answers begin with the same letters as the characters you have revealed so far. Do you understand? Output Yes or No, just that.}
    \item Prompt explaining the rules to the guesser:\\\\\texttt{You are playing a game with other guesser players against a setter player who is slowly revealing the initial letters of the word. The setter player will initially tell you just the first letter of the word. Each round, you can find a random word that starts with the initial letters revealed so far. Then, you need to come up with a meaningful clue or a description of this word and reveal it to other guesser players. You are not allowed to have a clue that is very similar to the word itself. If the word you found is not the same as the word that the setter came up with, the setter will try to guess your word and block it by saying your word. If some other guesser player can correctly guess your word, then the setter player will reveal one more letter. If the word both the guesser players guessed is the same as the word the setter player came up with, you all win. In every round, you can either choose to make a clue or try to guess from some other guesser player's clue. Note that in every round, your word must start with the initial letters revealed so far. Do you understand? Output Yes or No, just that}
    \item Prompt to generate a word from a clue:\\\\\texttt{You have been given the clue \dots. Now, guess a single word that could be a possible answer to this clue, starting with the letters \dots. Make sure your word is NOT one of these words: \dots and is different. Just output this word, do not output anything else.}
    \item Prompt to generate a clue:\\\\\texttt{The partial word you know so far is\dots. Come up with a word that starts with \dots. Make sure your word is NOT one of these words: \dots and is different. Just output this word, do not output anything else.''}
    \item Prompts to correct an agent:\begin{itemize}
        \item \texttt{Your earlier word does not start with \dots. Try again. Come up with a word that starts with \dots. Make sure your word is NOT one of these words: \dots and is different. Just output this word, do not output anything else.}
        \item \texttt{Your earlier word does not start with \dots. Try again. You have been given the clue \dots. Now, guess a single word that could be a possible answer to this clue, starting with the letters \dots. Make sure your word is NOT one of these words: \dots and is different. Just output this word, do not output anything else.}
        \item \texttt{Your earlier word cannot be one of these words: \dots. Try again. Come up with a word that starts with \dots. Make sure your word is NOT one of these words: \dots and is different. Just output this word, do not output anything else.}
        \item \texttt{Your earlier word cannot be one of these words: \dots. Try again. You have been given the clue \dots. Now, guess a single word that could be a possible answer to this clue, starting with the letters \dots. Make sure your word is NOT one of these words: \dots and is different. Just output this word, do not output anything else.}
    \end{itemize}
\end{itemize}
\end{document}